# A framework with updateable joint images re-ranking for Person Re-identification

Yuan Mingyue[1,2]  Yin Dong[1,2*]  Ding Jingwen[1,3*]  Luo Yuhao[1,2]  Zhou Zhipeng[1,2]
Zhu Chengfeng[1,2]  Zhang Rui[1,2]

[1]School of Information Science Technology, USTC, Hefei, Anhui 230027, China
[2]Key Laboratory of Electromagnetic Space Information of CAS, Hefei, Anhui 230027, China

**Abstract** Person re-identification plays an important role in realistic video surveillance with increasing demand for public safety. In this paper, we propose a novel framework with rules of updating images for person re-identification in real-world surveillance system. First, Image Pool is generated by using mean-shift tracking method to automatically select video frame fragments of the target person. Second, features extracted from Image Pool by convolutional network work together to re-rank original ranking list of the main image and matching results will be generated. In addition, updating rules are designed for replacing images in Image Pool when a new image satiating with our updating critical formula in video system. These rules fall into two categories: if the new image is from the same camera as the previous updated image, it will replace one of assist images; otherwise, it will replace the main image directly. Experiments are conduced on Market-1501, iLIDS-VID and PRID-2011 and our ITSD datasets to validate that our framework outperforms on rank-1 accuracy and mAP for person re-identification. Furthermore, the update ability of our framework provides consistently remarkable accuracy rate in real-world surveillance system.
**Keywords** person re-identification, video surveillance, image pool, convolutional network

# 1. Introduction

Recognizing the same object is an important task in computer vision [1]. In recent years, person re-identification (re-id) has been developed from object match and recognition and become an important branch. The prime target of person re-id is to identify the same person in a cross-camera under different conditions [2]. Accurate re-id is crucial for robust wide-area tracking where persons are tracked as they move through a camera-network, and may be useful for single-camera tracking [3]. Due to large appearance changes caused by environmental and geometric variations as a person moves among cameras, person re-id is still a challenging task.

In general, person re-id works mainly have two classes, image-based and video-based [1]. In traditional image-based re-id research, matching two cropped pedestrian images is the basic method to recognize a person in different cameras [4]. Furthermore, the way of calculating distance between the probe image and gallery images is also a critical step to impact its accuracy. A number of works about metric learning [5, 6, 7, 8] and re-ranking method [9, 10] are utilized to improve accuracy. Video-based person re-id works use sequences of person's video frames rather than only a single image. Traditional methods extract pose message from image sequences [11]. McLaughlin et al. [3] first applies deep learning to the video re-id problem in 2016. [12] develops and uses the Long-Short Term Memory (LSTM) network to aggregate frame-wise person features in a recurrent manner. Even though video-based methods effectively improve over the single-frame methods and the re-id accuracy will saturate as the number of selected frames increases [13], video-based methods still have limitations for multi-camera systems. Back to the begin of the person re-id, its first purpose is to realize multi-camera tracking [14], while the current re-id works mainly concerned with two cameras [1].

In real applications, such as the video surveillance system of airport, railway station, campus and the mall. There are many cameras widely across entire conditions. The person re-id task needs to track the object through all cameras continually. While in traditional person re-id works, both image-based and video-based only use images from one camera as probe. Other information from multiple cameras is wasted [15]. The number of cameras that identified the target person is increasing over time in the entire surveillance system, therefore the diversity of stored images about the target person will also be improved. Furthermore, the re-id accuracy will enhance with the growth of the diversity of the target [15]. To overcome the weakness of the image-based and video-based methods for multi-camera tracking in practical complex condition, we combine re-ranking method and multiple images method and propose our solution: a novel framework, which uses a reservoir space: Image Pool to store images of the target person from different cameras in whole system. Then we use those images together to identify the target person and update images in Image Pool adaptively.

The contribution of this paper is three-fold. (1) We use mean-shift tracking method to select video frames of the target person automatically. Those video frames compose a collection called Image Pool. We propose a re-ranking framework, which can re-rank the original ranking list of the main image by calculating whole images in Image Pool. (2) We propose new rules for updating images in Image Pool. When the whole surveillance system keeps tracking the target person, a true new sample satisfied with our updating critical formula can be added to Image Pool following updating rules. It is worth mentioning that there are two situations of updating rules, the new image is from the current camera and from a disjoint camera. (3) Since the standard datasets of person re-id are only collected from two cameras, it is impossible to verify the update ability of our framework. We collect and create a new person re-id dataset called indoor train station dataset (ITSD). We validate that our updateable framework can provide continually remarkable accuracy rate and compare our framework with the state of the art on Market-1501, iLIDS-VID and PRID-2011 datasets.

## 2. Related work

### 2.1 Feature extraction

Conventional works on person re-id mainly focus on the invariant feature representation and distance metric learning. Discriminative features that are invariant to environmental and view-point changes incontrovertible play a determining role in the result of person re-id performance. [1] combines spatial and color information using an ensemble of discriminant localized features and classifiers to improve viewpoint invariance. Supervised learning based methods that map the raw features into a new space have greater discriminative power [16, 17, 21].

The current deep learning networks have led to a series of breakthroughs and greatly increase recognition accuracy than tradition ways. The depth of deep learning network has a great impact on the effect identification task. The "levels" of features can be enriched by the number of stacked layers (depth)[18]. However, there is a degradation problem: with the network depth increasing, accuracy gets saturated and then degrades rapidly. In this case, we use residual learning framework (ResNet) to address the degradation problem. The principle of ResNet is that to fit layers with a residual mapping rather than fit each few stacked layers directly with a desired underlying mapping.

The researchers of ResNet present successfully trained models on this dataset with over 100 layers, and explore models with over 1000 layers. ResNet won the 1st place on the ImageNet classification dataset [19] in the ILSVRC 2015 classification competition. ResNet obtains excellent results by extremely deep residual nets and still has lower complexity than VGG nets [20] and has 3.57% top-5 error on the ImageNet test set.

## 2.2 Distance calculation

Distance metric learning is used in person re-id to emphasize inter-person distance and deemphasize intra-person distance by calculating distance between feature vectors. Various methods have been proposed such as, Relaxed Pairwise Learning (PRLM) [17], Large Margin Nearest-Neighbour (LMNN) [21], and Relevance Component Analysis (RCA) [22].

After the comparison of experiments, we use Euclidean to calculate distance between the feature vectors of two images in our actual person re-id system. Given a pair of images $(s_i, s_j)$, each sequence feature vectors, $v_i = R(s_i)$ and $v_j = R(s_j)$ were extracted from images by the neural network [3]. We can write network training objective as a function of the feature vectors $v_i$ and $v_j$ as follows:

$$E(v_i, v_j) = \begin{cases} \frac{1}{2} \|v_i - v_j\|^2 & i = j \\ \frac{1}{2} \left[\max\left(m - \|v_i - v_j\|\right), 0\right]^2 & i \neq j \end{cases} \quad (1)$$

$\|v_i - v_j\|$ is the Euclidean distance between two images. When the sequences are from the same person, $i = j$, the objective encourages the features $v_i$ and $v_j$ to be close, as measured by Euclidean distance, while for images from different persons, $i \neq j$, the objective encourages the features to be separated by a margin $m$.

### 2.3 Re-ranking method

Re-ranking method is a step which can receive higher ranks with initial ranking list and the relevant images. Re-ranking methods have been studied to improve object retrieval accuracy in some works [23]. Shen et al. [24] utilize the k-nearest neighbors method to explore similarity relationships to achieve the re-ranking goal. The new score of each image is calculated depending on its positions in the produced ranking lists. Chum et al. [25] propose the average query expansion method, where a new query vector is obtained by averaging the vectors in the top-k returned results, and is used to re-query the database. To take advantage of the negative samples which have considerable distance with the query image, Arandjelovic et al. [26] develop the query expansion by using a linear SVM to obtain a weight vector. The distance from the decision boundary is employed to revise the initial ranking list.

### 2.4 Multiple images for re-id

Multiple images for re-id is a method to achieve person re-id by multiple images rather than a single image. The use of multiple images can be exploited to improve performance in many realistic scenarios. Existing methods for multi-shot re-id include collecting interest-point descriptors over time [27], or training classifiers using features collected over multiple frames [28]. In addition, supervised learning based methods have also been used, such as learning a distance preserving low-dimensional manifold [29], or learning to map among the appearances in sequences by taking into account the differences between specific camera pairs [30]. Furthermore, there are some approaches that explicitly model video include using a conditional random field (CRF) to ensure similar images in a video sequence receive similar labels [31], or extracting space-time features [32] and learning a ranking function that is robust to partially corrupted sequences [33].

## 3. Method

### 3.1 Multiple-Image joint distance

In this subsection, we illustrate the algorithms of multiple-image joint distance. We continuously collect images of the target person into a set in our framework, and we call it Image Pool (Rules of updating Image Pool will be introduced in the part of 3.3). Given a probe person Image Pool $\Omega$ (showed in Table 1) with

$M$ images $\Omega_{ID} = \{\omega_i | i = 1, 2, \ldots, M\}$, and the gallery set $G$ with $N$ images $G = \{g_j | j = 1, 2, \ldots, N\}$. The distances $S(g_j, \Omega_{ID})$ between each image $g_i$ in the G and all the images in the Image Pool need to be calculated. Depending on $S(g_j, \Omega_{ID})$, we can rank images in the gallery. In the result, the higher the rank of the image represents the greater the similarity between the image and the target person. The ranking list of gallery images is $R(G, \Omega_{ID}) = \{g_1^1, g_2^1, g_3^1, \ldots, g_N^1\}$, where $S(g_j^1, \Omega_{ID}) < S(g_{j+1}^1, \Omega_{ID})$.

| Image Pool | | | | | | | |
|---|---|---|---|---|---|---|---|
| Camera1 | Camera2 | Camera3 | Camera4 | Camera5 | Camera6 | Camera7 | Camera8 |
| 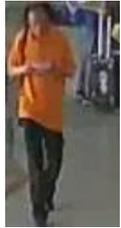 | 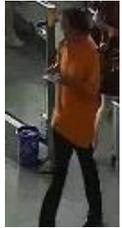 | 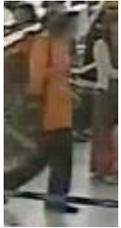 | 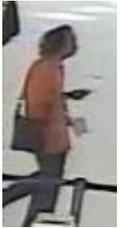 | 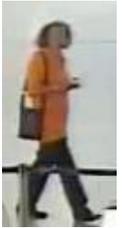 | 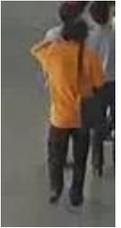 | 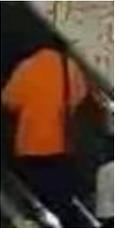 | 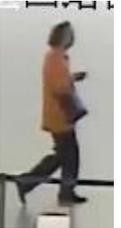 |
| I ($n_1$, 1) | I ($n_2$, 4) | I ($n_3$, 7) | I ($n_4$, 10) | I ($n_5$, 13) | I ($n_6$, 16) | I ($n_7$, 19) | I ($n_8$, 22) |
| 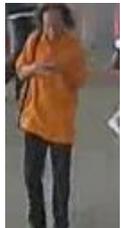 | 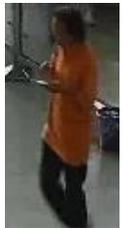 | 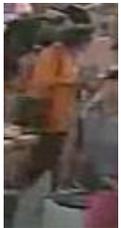 | 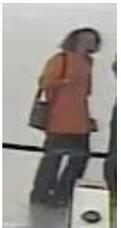 | 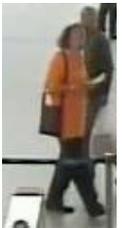 | 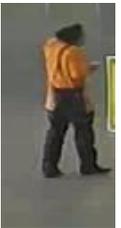 | 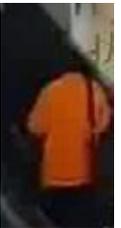 | 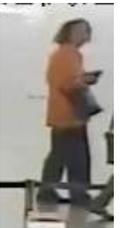 |
| I ($n_1$, 2) | I ($n_2$, 5) | I ($n_3$, 8) | I ($n_4$, 11) | I ($n_5$, 14) | I ($n_6$, 17) | I ($n_7$, 20) | I ($n_8$, 23) |
| 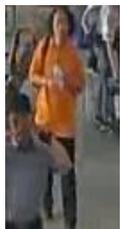 | 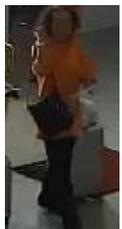 | 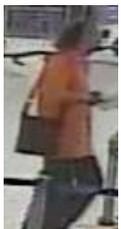 | 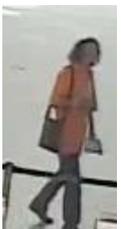 | 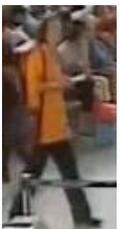 | 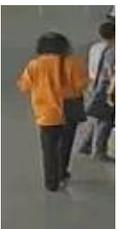 | 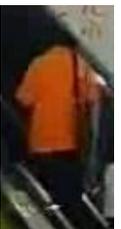 | 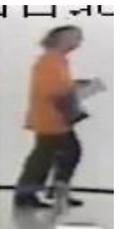 |
| I ($n_1$, 3) | I ($n_2$, 6) | I ($n_3$, 9) | I ($n_4$, 12) | I ($n_5$, 15) | I ($n_6$, 18) | I ($n_7$, 21) | I ($n_8$, 24) |

Table 1. Image Pool. Shown is an examplar set for one person. All these images in the users personal photo collection from different cameras are clustered together. For I($n_x$, f), $n_x$ represents that the current image is from the x camera and f is the number of the image in Image Pool.

We use $f(g_j, \omega_i)$ to represent similarity and define a map function from

Euclidean distance to similarity as Eq.2.

$$f(g_j, \omega_i) = \begin{cases} \frac{1}{E(g_j, \omega_i)} & E(g_j, \omega_i) \leq \kappa \\ 0 & E(g_j, \omega_i) > \kappa \end{cases} \quad (2)$$

The Euclidean distance of some images is too large and those images are obviously not the target people in the calculations. Therefore, the distance threshold $\kappa$ is set to reduce computation.

We choose an image $\omega_i$ in Image Pool $\Omega$ and rename it with $\omega_{main}$, set it with the maximum weight value. The Multiple-Image joint distance function between gallery image $g_j$ and Image Pool $\Omega_{ID}$ can be written as follow:

$$S(g_j, \Omega_{ID}) = E(g_j, \omega_{main}) - \sum_{i=1}^{M}\left[ W_i \cdot \frac{\eta f(g_j, \omega_i)}{\sum_{i=1}^{M} f(g_j, \omega_i)} \cdot E(g_j, \omega_{main}) \right] \quad (3)$$

Where $E(g_j, \omega_{main})$ is the Euclidean distance between main image and a gallery image $g_j$. $\eta$ is a scale factor and $W_i (i=1,2,\ldots,M-1)$ are the weights of each image in Image Pool. $\frac{\eta f(g_j, \omega_i)}{\sum_{i=1}^{M} f(g_j, \omega_i)} \cdot E(g_j, \omega_{main})$ is used as normalization of $f(g_j, \omega_i)$. The lower value of $S(g_j, \Omega_{ID})$ indicates that the image $g_i$ is more likely the target person.

There remains some aspects can be improved for our Multiple-Image joint distance algorithm: (1) It is time-consuming to calculate $S(g_j, \Omega_{ID})$ and all image pairs between sets gallery G and Image Pool $\Omega$. An alternative way is to re-rank the initial ranking list of Euclidean distance. (2) It is difficult to determine accurately suitable parameter of Eq.2 and Eq.3, such as scale factor $\eta$, and threshold $\kappa$.

### 3.2 Multiple-Image joint re-ranking

In order to improve the aspects mentioned above, we proposed Multiple-Image joint re-ranking algorithm based on Multiple-Image joint distance algorithm. We still choose an image $\omega_{main}$ in Image Pool $\Omega$. We get initial main ranking list $R(G,\omega_{main}) = \{g_1^1, g_2^1, g_3^1, \ldots, g_N^1\}$ by ascending order Euclidean distance, which is calculated between probe $\omega_{main}$ and a gallery image $g_j (j=1,2,\ldots,N)$. Accordingly, the ranking list of other images in Image Pool is:

$$R(G, \omega_i) = \{g_1^1, g_2^1, g_3^1, \ldots, g_N^1\} \ (i = 1,2, \ldots, M - 1) \tag{4}$$

We take top-k samples of the ranking list to re-rank. The top-k ranking list is defined as follow:

$$R_k(G, \omega_i) = \{g_1^1, g_2^1, g_3^1, \ldots, g_k^1\} \ (i = main, 1, 2, \ldots, M - 1) \tag{5}$$

If the ranking list of an image in $\Omega$ is more reliable, we set the weight of the image with a bigger value. The number of candidate $k_i$ in main image ranking list should be the maximum. In order to simplify the calculation, the sum of all weights is 1. Main image weight is 0.5. The rests of images in $\Omega$ have the same weights. The weights are defined as follow:

$$W_i = \begin{cases} 0.5 & i = main \\ \dfrac{1}{2(M-1)} & i = 1,2,3, \ldots, M-1 \end{cases} \tag{6}$$

the $k_i$ is defined as follow:

$$k_i = \left[ \eta \cdot \dfrac{W_i}{\sum\limits_{i=0}^{M-1} W_i} \right] \ (i = main, 1, 2, \ldots M - 1) \tag{7}$$

$\eta$ is a scale factor. $[\cdot]$ is a rounding operation. The algorithm of Multiple-Image joint re-ranking is showed in Algorithm 1 and the illustration of Multiple-Image joint re-ranking algorithm of a person re-id application is showed in Figure 1.

---

**Algorithm 1**: Multiple-Image joint re-ranking

Input: $R_{k1}(G,\omega_{main}) = \{g_1^1, g_2^1, \ldots, g_{k1}^1\}$  $R_{k2}(G,\omega_i) = \{g_1^1, g_2^1, \ldots, g_{k2}^1\}(i=1,2,\ldots,M-1)$

Begin

   for $g = g_1^1 : g_{k1}^1$ do

```
    for m = g_i^1 (g_i^1 ∈ R_{k2}(G, ω_i)) do
       1. If (g= =m)
          Count g belong to R_{k2} times T.
          Generate the re-ranking list with image g by descending order of T.
       2. If there is an additional second main image in Image Pool, the rest of
          images in Ω do not have the same weights. If g both belongs to
          R_{k1}(G, ω_i) and R_{k2}(G, ω_i), we will top it in re-ranking list.
       end do
       Initialize T to 0.
    end do
end
Output: R_{k1}^*(G, Ω) = {g_1^2, g_2^2, g_3^2, …, g_{k1}^2}
```

Additional condition: if the conditions of two cameras are very similar, such as light, camera's perspective and other environmental conditions, we set the image under this camera as the second main image.

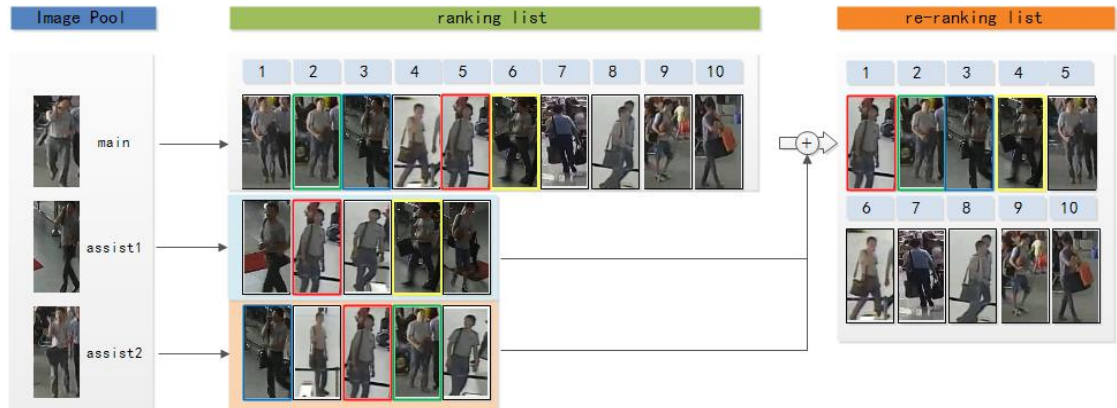

Figure 1. **Illustration of Multiple-Image joint re-ranking algorithm of a person re-id application.** Ranking list: The query of images in Image Pool and its K-nearest neighbors, where k1 is 10, k2 is 5. The same color box images are the same picture. Red box image is included in all ranking list, and green, blue, yellow is included 2 ranking list respectively. Re-ranking list: gained by re-ranking the list of main image. Red box image is top-1 and other color boxes are brought forward to front of list.

### 3.3 Rules of updating Image Pool

We declare the algorithm of the Multiple-Image joint distance between gallery image $g_j$ and Image Pool $Ω$ in 3.1. Furthermore, we propose Multiple-Image joint re-ranking algorithm in 3.2. Whether the images in the Image Pool can fully express the target person, person re-id accuracy will be affected. Consequently, we propose the rules of updating Image Pool. With those rules, we can update Image Pool in practical surveillance system automatically. The rules of updating Image Pool are showed in Algorithm 2 and the illustration and application of rules are showed in Figure 2. $I(n,f)$ represents that the image $I(n,f)$ is from number $n$ camera and frame $f$.

---

**Algorithm 2:** The rules of updating Image Pool

---

**Input:** videos from different cameras in whole surveillance system.
Begin
    1. **Initialization:** we select the target person with a rectangular box by hand, which image is marked as $I(n_1, f_1)$. We use [37] tracking method to get other $M-1$ images every $\beta$ frames. Eventually, we get initial Image Pool $\Omega = \{I(n_1, f_1), I(n_1, f_2), \ldots, I(n_1, f_M)\}$.

- **Updating (current camera):** when system gets new correct image $I(n_1, f_{new})$ of target person in current camera, we calculate the mean of the distance between $I(n_1, f_{new})$ and all the images in the $\Omega$. If the result of calculation fits Eq. 8, then:

    Add $I(n_1, f_{new})$ in $\Omega$.

    Find the image with maximum value of $E(g_{new}, g_n)\ \ n \in \Omega$ and copy $W$ of this image to new image.

    Delete the image with maximum value of $E(g_{new}, g_n)$.

- **Updating (cross camera):** when system gets new correct image $I(n_2, f_{new})$ of target person in cross camera, then:

    Add $I(n_2, f_{new})$ in $\Omega$.

    Copy the value of $W$ and picture number from the main image to new image.

    Find the image with maximum value of $E(g_{new}, g_n)\ \ n \in \Omega$ and copy $W$ of this image to main image.

    Delete the image with maximum value of $E(g_{new}, g_n)$.

End

**Output:** new Image Pool $\Omega^* = \{I(n_1^*, f_1^*), I(n_1^*, f_2^*), \ldots, I(n_1^*, f_M^*)\}$

In Eq. 8, we judge that whether the mean of the distance between $I(n_1, f_{new})$ and all the images in the $\Omega$ is larger than threshold $\gamma$.

$$\frac{1}{k}\sum_{n=1}^{k} E(g_{new}, g_n) > \gamma \tag{8}$$

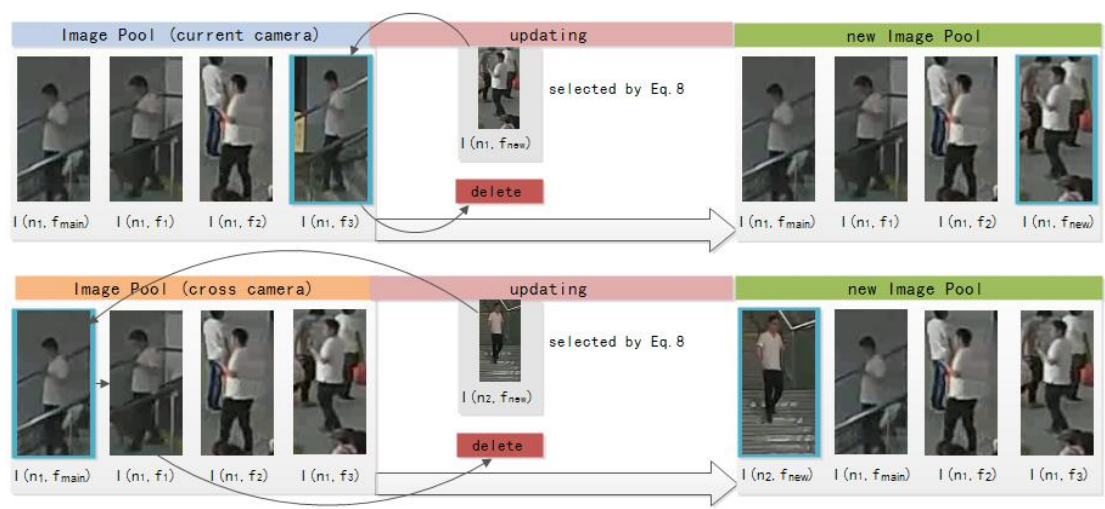

Figure 2. **The rules of updating Image Pool of a person application.** Figure shows the update process of the Image Pool. Top: new image is from the same camera with previous one. When this image is satisfied with Eq. 8, we use it to replace the most dissimilar picture in Image Pool, then new Image Pool is generated. Bottom: new image is from a disjoint camera with previous one, which is satisfied with Eq. 8. We generate new Image Pool by new image taking over main image and delete the most dissimilar image in Image Pool.

### 3.4 Joint re-ranking framework

Our Joint re-ranking framework is illustrated in Figure 3. Given a gallery set ($N$ images) and an Image Pool, which contains $M$ images. The feature vectors are extracted for each person by convolutional neural network. Then the distances are calculated for each pair of the probe image and gallery image. Ranking lists are obtained by ascending order Euclidean distance. The re-ranking list is obtained by the Multiple-Image joint re-ranking algorithm.

In addition, we design a system that can generate Image Pool and achieve person re-id in real video surveillance system. The system is showed in Figure 4. The system is especially designed for our ITSD dataset. There are 16 cameras in the entire station video surveillance system. When we first calibrate the target person, this system can generate Image Pool automatically by mean-shift tracking method. When our system keeps tracking the target throughout the video system, it updates Image

Pool with new positive images following our updating rules continuously. Final, we achieve continually effective person re-id in real surveillance application.

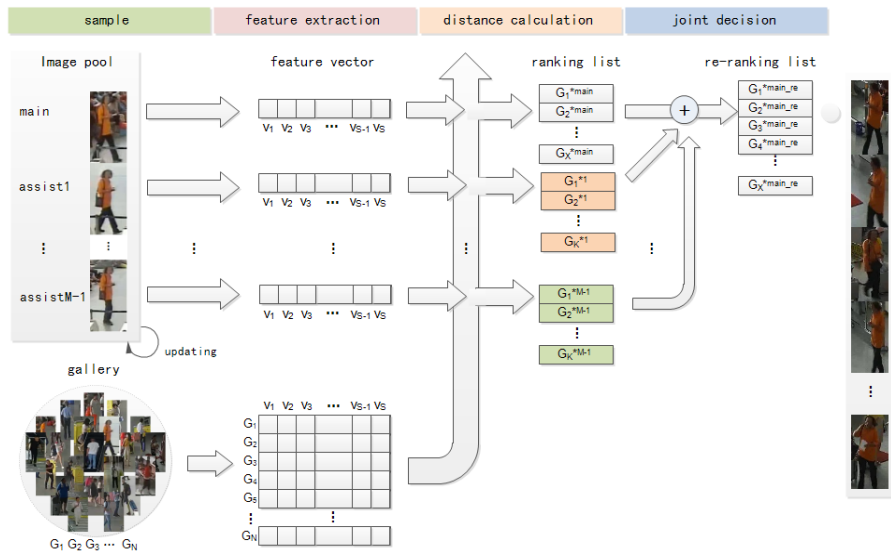

Figure 3. Proposed joint re-ranking framework for person re-id.

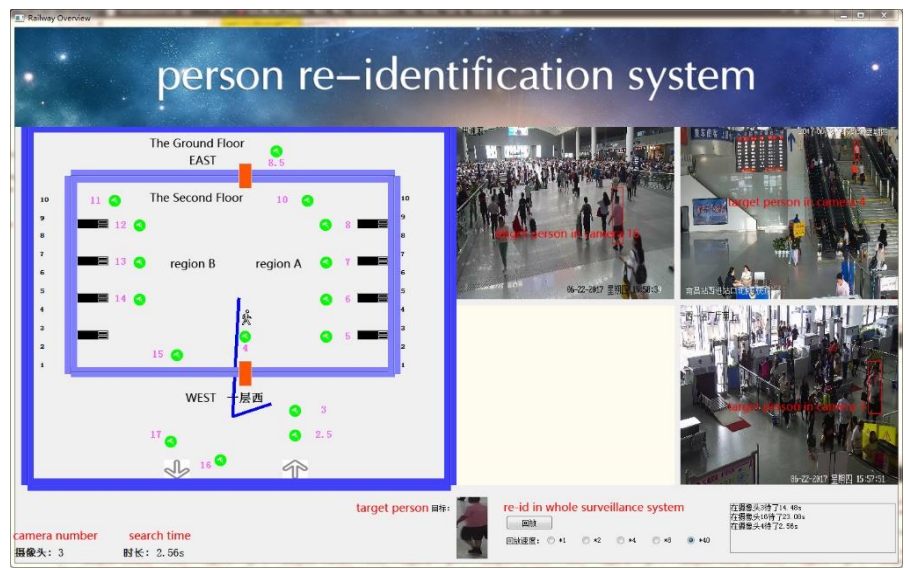

Figure 4. Designed person re-id system in real video surveillance system. Figure shows that this target person is tracked by three cameras in whole surveillance system. The left part of the figure shows the plane map of the railway station and the location of each camera. The right part of the figure shows the result of recognizing the target by the camera 4, 16, 3 in turn.

## 4. Experiment

### 4.1. Dataset

Since our framework is based on multiple images, we conducted experiments on four datasets. Two image sequence datasets: the PRID 2011 dataset [16], iLIDS-VID dataset [33]. One image-based dataset Market-1501 [34] and our newly dataset: Indoor train station dataset (ITSD).

**PRID 2011 dataset** – The PRID 2011 re-id dataset [16] includes 400 image sequences for 200 people from two camera views that are adjacent to each other. Each image sequence has variable length consisting of 5 to 675 image frames.

**iLIDS-VID dataset** – iLIDS-VID person sequence dataset [33] has been created based on two non-overlapping camera views. It consists of 600 image sequences for 300 randomly sampled people, with one pair of image sequences from two camera views for each person. Each image sequence has variable length consisting of 23 to 192 image frames.

**Market-1501** – Market-1501 [34] is currently the largest image-based re-ID benchmark dataset. It contains 32,668 labeled bounding boxes of 1,501 identities captured from 6 different viewpoints.

**ITSD dataset** – In order to verify the update ability of our framework, we collect and annotate a small indoor train station dataset. It contains 5607 images, 443 identities captured from 8 different viewpoints. 50 identities are captured by at least 3 disjoint cameras. The images are detected by DPM method and selected by using supervised mean-shift tracking method.

### 4.2. Experimental settings and Evaluation protocol

**Experimental settings**: for iLIDS-VID and PRID 2011 datasets, the total pool of sequence pairs is randomly split into two subsets of equal size, one for training and one for testing. Following the evaluation protocol on the PRID 2011 dataset [33], in the testing phase, the sequences from the first camera are used as the probe set while the ones from the other camera as the gallery set. For Market-1501, the dataset is split into two parts: 12,936 images with 751 identities for training and 19,732 images with 750 identities for testing. In testing, 3,368 hand-drawn images with 750 identities are used as probe set to identify the correct identities on the testing set. For ITSD dataset, the dataset is split into two parts: 3833 images with each identity for training and 1774 images with each identity for testing.

Training is all implemented by using the Caffe framework on NVIDIA GeForce GTX 1080 GPU. We use ResNet-50 [18] network to extract features. The network was trained for 50000 epochs using stochastic gradient descent with a learning rate of 1e-3, and a batch size of 16.

**Evaluation protocol:** In experiments, we use two evaluation metrics to evaluate the performance of re-ID methods on all datasets. The first one is the Cumulated Matching Characteristics (CMC). Considering re-ID as a ranking problem, we report the cumulated matching accuracy at rank-1. The other one is the mean average precision (mAP). Considering re-id as an object retrieval problem, it was described in [34]. For image-based datasets, the results of Image Pool are considered as the results of main image. And for image sequence datasets, the results of Image Pool are considered as the results of whole sequence.

### 4.3. Parameters Analysis

The parameters of our framework are analyzed in this subsection. The feature extractor is effectively trained on classification model including CaffeNet [35] and ResNet-50 [18]. CaffeNet generates a 1,024-dim vector and ResNet-50 generates a 2,048-dim vector for each image. We evaluate the influence of *k1*, *k2*, and *M* of Image Pool on rank-1 accuracy and mAP on the Market-1501 dataset. The results of Image Pool are considered as the results of main image. In experiments, every image in probe set is used as main image for once. Other images in Image Pool are selected by our rules (current camera). It is fair to compare the results with other methods.

We set k1 to 10, k2 to 10. Moreover, experiments conducted with three metrics, Euclidean, KISSME [36] and XQDA [5] verify the effectiveness of our method on different distance metrics. Results in Table 2 show the influence of changing the number of *M*. In the results, both the rank-1 accuracy and mAP are significantly improved with our framework in all experiments. The rank-1 increases with the growth of *M*. The mAP first increases with the growth of *M*, then begins a slow decline after *M* surpasses 3. Time in Table 2 is the total calculation time of 3368 query images. The maximum value of average time of every query is 0.03517s and the minimum value is 0.01613s in experiments. When the *M* is set to 3, we can minimize the calculation time. In general, ResNet-50 model performance is better than CaffeNet. XQDA metrics has best performance on mAP and calculation time, while Euclidean outperforms in rank-1 accuracy.

Table 2. The influence of the number of images (parameter M) in Image Pool. The k1 is fixed at 10 and k2 is fixed at 10.

| Method | Rank-1 | mAP | Time |
| --- | --- | --- | --- |
| CaffeNet+Euclidean | 56.98 | 31.41 | 94.035464 |
| CaffeNet+Euclidean+joint(M=2) | 67.76 | 35.17 | 118.458838 |
| CaffeNet+Euclidean+joint(M=3) | 74.44 | 35.90 | 109.906106 |
| CaffeNet+Euclidean+joint(M=4) | 75.33 | 35.78 | 112.739676 |
| ResNet-50+XQDA | 73.81 | 51.19 | 93.118241 |
| ResNet-50+XQDA+joint(M=2) | 80.97 | 54.79 | 78.275997 |
| ResNet-50+XQDA+joint(M=3) | 83.40 | 54.26 | 54.338506 |
| ResNet-50+XQDA+joint(M=4) | 84.20 | 54.19 | 56.466179 |
| ResNet-50+KISSME | 75.68 | 50.95 | 92.781606 |
| ResNet-50+KISSME+joint(M=2) | 82.51 | 52.27 | 84.042617 |
| ResNet-50+KISSME+joint(M=3) | 84.38 | 54.06 | 64.588775 |
| ResNet-50+KISSME+joint(M=4) | 85.84 | 54.10 | 66.142786 |
| ResNet-50+Euclidean | 73.28 | 47.87 | 82.399024 |
| ResNet-50+Euclidean+joint(M=2) | 81.77 | 52.27 | 84.042617 |
| ResNet-50+Euclidean+joint(M=3) | 85.15 | 51.62 | 62.396900 |
| ResNet-50+Euclidean+joint(M=4) | 86.05 | 51.60 | 62.425897 |

With the best performance of our approach in Table 2: ResNet-50+Euclidean+joint(M=3), we evaluate the influence of $k1$, $k2$, and on rank-1 accuracy and mAP on the Market-1501 dataset.

In Table 3, $k2$ is fixed to 6. The mAP increases with the growth of $k1$. When $k1$ grows, the rank-1 accuracy first rises, and after arriving at the optimal point around $k1 = 70$, it starts to drop. The reason of decline in performance is that too large value of $k1$ causes more false matches.

In Table 4, $k1$ is fixed to 70. The mAP first increases with the growth of $k2$, and then begins a slow decline after $k2$ surpasses the value of 8. The rank-1 accuracy reduces with the growth of $k2$ (when $k2$ is greater than 2). The reason of decline in performance is that the large value of $k2$ causes more false matches and significantly impacts the rank-1 accuracy.

Table 3. The impact of the parameter $k1$ on re-ID performance on the Market-1501. The baseline method is ResNet-50+Euclidean+joint(M=3). The $k2$ is fixed to 6.

| Method | Rank1 | mAP |
|---|---|---|
| k1=6, k2=6 | 82.24 | 50.69 |
| k1=10, k2=6 | 85.72 | 52.07 |
| k1=20, k2=6 | 88.90 | 52.66 |
| k1=30, k2=6 | 89.85 | 53.53 |
| k1=40, k2=6 | 90.56 | 57.79 |
| k1=50, k2=6 | 90.88 | 58.80 |
| k1=60, k2=6 | 91.09 | 59.45 |
| **k1=70, k2=6** | **91.75** | **59.31** |
| k1=80, k2=6 | 91.48 | 60.59 |
| k1=90, k2=6 | 91.45 | 61.00 |
| k1=100, k2=6 | 91.54 | 61.33 |

Table 4. The impact of the parameter $k2$ on re-ID performance on the Market-1501. The baseline method is ResNet-50+Euclidean+joint(M=3). The $k1$ is fixed to 70.

| Method | Rank1 | mAP |
|---|---|---|
| k1=70, k2=20 | 87.25 | 58.84 |
| k1=70, k2=15 | 87.68 | 59.62 |
| k1=70, k2=10 | 89.40 | 60.20 |
| k1=70, k2=9 | 89.90 | 60.26 |
| k1=70, k2=8 | 90.44 | 60.29 |
| k1=70, k2=7 | 90.97 | 60.29 |
| k1=70, k2=6 | 91.39 | 60.12 |
| k1=70, k2=5 | 92.04 | 59.93 |
| k1=70, k2=4 | 93.11 | 69.67 |
| k1=70, k2=3 | 93.91 | 59.03 |
| **k1=70, k2=2** | **94.80** | **57.60** |
| k1=70, k2=1 | 94.51 | 54.73 |

### 4.3 Rules of updating Image Pool

In this subsection, we evaluate the rules of updating Image Pool of our framework on our ITSD dataset with the best performance of our approach:

ResNet-50+Euclidean+joint(M=3), *k1* is set to 70 and *k2* is set to 2. The results of Image Pool are considered as the results of main image. Other images in Image Pool are generated by mean-shift tracking method, while in the experiments, they selected by different rules.

**Table 5.** The performances of Rank CMC(%) and mAP of selecting assist images by different rules. The baseline method is ResNet-50+Euclidean, and the results are from every single main image. (a) The assist images is selected from the same camera as the main image randomly. (b) The assist images is selected from the same camera as the main image by our rules. (c) The assist images is selected from both current camera and cross camera randomly. (d) The assist images is selected from cross camera by our rules.

| Method | Rank-1 | Rank-5 | Rank-10 | Rank-20 | mAP |
|---|---|---|---|---|---|
| baseline | 37.92 | 60.10 | 67.85 | 74.72 | 41.63 |
| (a) | 44.19 | 61.79 | 64.00 | 73.19 | 46.75 |
| (b) | 54.79 | 69.86 | 69.30 | 73.48 | 55.09 |
| (c) | 58.01 | 73.76 | 75.34 | 77.48 | 60.27 |
| (d) | 58.53 | 78.17 | 81.44 | 85.00 | 64.50 |

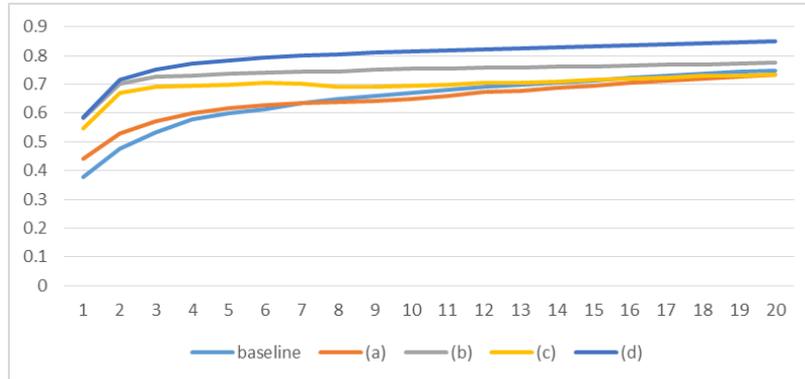

**Figure 5.** CMC curves comparison with different Image Pool rules.

From Table 5 and Figure 5, compare to baseline, our framework improves re-id accuracy evidently. Compare to the results of 5(b) and 5(a), we can see that selecting image by our updating rules (current camera) is much effective than selecting images randomly. Comparing the results of 5(d) to 5(c), it can also illustrate that selecting images by our updating rules (cross camera) have good performance than selecting images from both camera randomly. Furthermore, comparing the results of 5(d) to 5(b), the rank-1 accuracy improves 3.74%, mAP improves 9.41%. It can be seen that, updating rules that are divided into cross camera and current camera two situation are effective and necessary.

### 4.4. Comparison with the state of the art

We now compare the performance of our proposed framework against state-of-art methods on large-scale person re-id benchmark datasets. To ensure a fair comparison, the experiments for our framework were trained and tested using the same datasets and same test/training split by following the evaluation protocol [33][34].

**Experiments on Market-1501**

We first compare our proposed framework on the largest image-based re-ID dataset. In Table 6, we compare the performance of our approach, ResNet-50+Euclidean+joint (M=3, k1=70, k2=2), with other state-of-the-art methods. Our best method impressively outperforms the previous work and achieves large margin advances compared with the state-of-the-art results in rank-1 accuracy and mAP.

Table 6. Comparison of our framework with state-of-the art on the Market-1501 dataset.

| Method | Rank-1 | mAP |
|:---:|:---:|:---:|
| Bow+Kissme [34] | 44.42 | 20.76 |
| SCSP [38] | 51.90 | 26.35 |
| Null Space [7] | 55.43 | 29.87 |
| LSTM Siamese [39] | 61.6 | 35.3 |
| Gated Siamese [40] | 65.88 | 39.55 |
| PIE [41] | 79.33 | 55.95 |
| SVDNet [42] | 80.5 | 55.9 |
| TinNet [43] | 84.92 | 69.14 |
| ours | 94.80 | 57.60 |

**Experiments on PRID-2011 and iLIDS-VID**

Results on PRID-2011 and iLIDS-VID: Comparing the Rank-1 accuracy and mAP shown in Table 6, we can see that our framework can achieve higher performance than all the compared methods for both PRID-2011 and iLIDS-VID datasets. The performance is boosted, especially for the Rank-1 protocol. The improvements are 3.2% and 1.7% for PRID-2011 and iLIDS-VID datasets respectively.

Table 6. Comparison of our framework with state-of-the art on PRID-2011 and iLIDS-VID.

| dataset | PRID-2011 | iLIDS-VID |
|---|---|---|
| Method | Rank-1 | Rank-1 |
| VR [33] | 42 | 35 |
| AFDA [44] | 43 | 38 |
| STA [45] | 64 | 44 |
| RFA [12] | 64 | 49 |
| RNN-CNN [3] | 70 | 58 |
| ASTPN [46] | 77 | 62 |
| **ours** | **81** | **66** |

## 5. Conclusion

In this paper we propose a novel framework for person re-id. We use mean-shift tracking method to select video frames of the target person automatically and those images compose a collection called Image Pool. We get original ranking list by calculating distance of feature vectors between Image Pool and gallery. Eventually, we get re-ranking list by following our Multiple-Image joint re-ranking algorithm. Furthermore, we propose new rules for updating images in Image Pool. When our system keeps tracking the target person throughout the video system, it updates Image Pool with new positive images following our updating rules continuously. Experiments validate that our updateable framework can provide continually remarkable accuracy rate both on three standard datasets and our ITSD dataset which bases on practical surveillance system. The further work of this paper is to test and improve our framework in more practical application scenarios.